\documentclass{article} 

\usepackage[letterpaper, left=1.2truein, right=1.2truein, top = 1.2truein, bottom = 1.2truein]{geometry}

\usepackage{amsmath}
\usepackage{amssymb}
\usepackage{mathtools}
\usepackage{amsthm}
\usepackage{amsfonts}
\usepackage{enumerate}
\usepackage[table]{xcolor}
\PassOptionsToPackage{pdftex}{graphicx}
\usepackage{tikz}
\usetikzlibrary{shapes.geometric, arrows}

\usepackage{subcaption}
\usepackage{multirow}

\tikzstyle{box} = [rectangle, rounded corners, 
minimum width=3cm, 
minimum height=.5cm,
text centered, 
draw=black]

\definecolor{lblue}{RGB}{40, 103, 178}
\definecolor{cred}{RGB}{177, 4, 14}
\definecolor{nyllw}{RGB}{245, 213, 71}

\usepackage{adjustbox}

\usepackage[blocks, affil-it]{authblk}

\usepackage[round]{natbib}

\usepackage{microtype}
\usepackage{booktabs} 

\usepackage[CJKbookmarks=true,
            bookmarksnumbered=true,  
			bookmarksopen=true,
			colorlinks=true,
			citecolor=blue,
			linkcolor=blue,
			anchorcolor=red,
			urlcolor=blue]{hyperref}

\usepackage[capitalize,noabbrev]{cleveref}
\Crefname{definition}{Defn.}{Defns.}
\Crefname{theorem}{Thm.}{Thms.}
\Crefname{prop}{Prop.}{Props.}

\newcommand{\R}{\mathbb{R}}

\newcommand{\III}[1]{{\left\vert\kern-0.25ex\left\vert\kern-0.25ex\left\vert #1 
    \right\vert\kern-0.25ex\right\vert\kern-0.25ex\right\vert}}

\newcommand{\calr}{\mathcal{R}} % only because \cr already taken
\newcommand{\ca}{\mathcal{A}}   \newcommand{\cd}{\mathcal{D}}                     

\newcommand{\eps}{\varepsilon}

\theoremstyle{plain}
\newtheorem{theorem}{Theorem}
\newtheorem{prop}[theorem]{Proposition}

\newtheorem{lemma}[theorem]{Lemma}

\newtheorem{definition}[theorem]{Definition}

\makeatletter
\providecommand*{\diff}%
	{\@ifnextchar^{\DIfF}{\DIfF^{}}}
\def\DIfF^#1{%
	\mathop{\mathrm{\mathstrut d}}%
		\nolimits^{#1}\gobblespace}
\def\gobblespace{%
	\futurelet\diffarg\opspace}
\def\opspace{%
	\let\DiffSpace\!%
	\ifx\diffarg(%
		\let\DiffSpace\relax
	\else
		\ifx\diffarg[%
			\let\DiffSpace\relax
	\else
		\ifx\diffarg\{%
			\let\DiffSpace\relax
		\fi\fi\fi\DiffSpace}

\newcommand\cR{\mathcal{R}}

\DeclareMathOperator{\argmax}{argmax}

\newcommand{\topk}{\textnormal{top-$k$}}
\DeclareMathOperator{\argsort}{ranking}
 \DeclareMathOperator{\ranking}{ranking}
\DeclareMathOperator{\dist}{dist}
\newcommand{\sargmax}{\argmax^{(\varepsilon)}}
\newcommand{\stopk}{\topk^{(\varepsilon)}}
\newcommand{\sranking}{\ranking^{(\varepsilon)}}

\usepackage{enumitem}
\setitemize{noitemsep,labelwidth=.5em,labelsep=.4em,topsep=0pt,parsep=0pt,partopsep=0pt,leftmargin=1em}

\usepackage[mathscr]{euscript}
 \let\mathscr\relax% just so we can load this and rsfs
\usepackage[scr]{rsfso}
\newcommand{\powerset}{\raisebox{.15\baselineskip}{\Large\ensuremath{\wp}}}

\AtBeginDocument{}

\title{Assumption-free stability for ranking problems}

\author[1]{Ruiting Liang}
\author[2]{Jake A. Soloff}
\author[2]{Rina Foygel Barber}
\author[2,3]{Rebecca Willett}
\affil[1]{Committee on Computational and Applied Mathematics, University of Chicago}
\affil[2]{Department of Statistics, University of Chicago}
\affil[3]{Department of Computer Science, University of Chicago}
\date{\today}

\begin{document}
\maketitle

\begin{abstract}
    In this work, we consider ranking problems among a finite set of candidates: for instance, selecting the top-$k$ items among a larger list of candidates or obtaining the full ranking of all items in the set. These problems are often unstable, in the sense that estimating a ranking from noisy data can exhibit high sensitivity to small perturbations. Concretely, if we use data to provide a score for each item (say, by aggregating preference data over a sample of users), then for two items with similar scores, small fluctuations in the data can alter the relative ranking of those items. Many existing theoretical results for ranking problems assume a separation condition to avoid this challenge, but real-world data often contains items whose scores are approximately tied, limiting the applicability of existing theory. To address this gap, we develop a new algorithmic stability framework for ranking problems, and propose two novel ranking operators for achieving stable ranking: the \emph{inflated top-$k$} for the top-$k$ selection problem and the \emph{inflated full ranking} for ranking the full list. To enable stability, each method allows for expressing some uncertainty in the output. For both of these two problems, our proposed methods provide guaranteed stability, with no assumptions on data distributions and no dependence on the total number of candidates to be ranked. Experiments on real-world data confirm that the proposed methods offer stability without compromising the informativeness of the output.

\end{abstract}

\section{Introduction}\label{sec:intro}
Ranking tasks arise across diverse machine learning settings, powering applications such as search, recommendation, and decision making. Despite their pervasiveness, instability remains a persistent challenge in many of the ranking algorithms employed in practice---the output ranking list is often sensitive to slight perturbations of the training data, which impairs their performance in real-world applications. For example, such sensitivity can sabotage selection fairness \citep{dwork2019learning, yang2018nutritional}, or degrade user experience in the context of recommendation systems \citep{anelli2021study, oh2022rank}.

In this paper, we develop a unified framework that addresses algorithmic stability in ranking problems by adopting a set-valued perspective on the output that allows for ambiguity in the ranking score assignments, such as near-tie scenarios. We propose methods to solve two problems, top-$k$ selection and full ranking, each leveraging recently developed tools in stable classification \citep{soloff2024building} to achieve provable ranking stability in an assumption-free way.

\subsection{Problem setting}\label{sec:problem-setting&prior-work}
In ranking tasks, we are given a list of $L$ items and a data-based mechanism for estimating the score of each item,
and we seek to either select the $k$ items with the top scores (the ``top-$k$ problem'') or rank the items according to their scores (the ``full ranking problem''). The challenge is that small perturbations in the data used to estimate scores can alter the selected set of $k$ items or the order of the ranked items.

Both problems are typically approached via a two-stage process: 
\begin{description}
    \item \textbf{Step 1:} given a dataset $\cd$ containing $n$ samples $Z_1,\dots,Z_n$, one runs an  algorithm $\ca$ to learn a vector of \textit{scores} $\hat{w}=(\hat{w}_1,\dots,\hat{w}_L)$ that assigns scores to the $L$ items;
    \item \textbf{Step 2:} given the scores contained in $\hat w$, one either sorts them (ranking) or selects the $k$ items with the largest scores (top-$k$ selection).
\end{description}
For example, for Step 1, in the learning-to-rank (LTR) problem \citep{liu2009learning,cao2007learning}, 
$\hat w$ represents the predicted relevance of each item to a query; this may be learned from training data $\cd$ using, for instance, a Bradley–Terry model \citep{bradley1952rank}.
As another example, in preference voting problems,  $\hat{w}_{\ell}$ denotes the fraction of votes the $\ell$-th item received in the dataset $\cd$, where each data value $Z_i$ indicates the choice of the $i$th voter.
In general, we  write $\hat{w} = \ca(\cd)$, where $\ca$ denotes the learning algorithm mapping a dataset $\cd=(Z_1,\dots,Z_n)$ to the score vector $\hat{w}$.
For Step 2, let 
$\pi$ denote the permutation on $[L]$ such that $\hat w_{\pi(j)}$ is the $j$-th largest element in $(\hat w_{1},\cdots,\hat w_{L})$ for each $j \in [L]$.
In top-$k$ selection, typically the $k$ elements with the largest scores are returned: $\topk(\hat w) := \{\pi(1),\ldots,\pi(k)\}$. 
For full ranking, a permutation over $[L]$ based on the sorted scores is produced, i.e., $\ranking(\hat w) := (\pi(1),\ldots,\pi(L))$. 
This two-stage process is summarized in the diagram below.

\begin{center}
\begin{tikzpicture}
\node (data) {\begin{tabular}{@{}c@{}}Training\\data\\$\cd\in\mathcal{Z}^n$\end{tabular}};
\node (alg) [box, fill=lblue!30, right of=data, xshift=.78in] {\begin{tabular}{@{}c@{}}Learning\\algorithm $\ca$\end{tabular}};
\node (weights) [right of=alg, xshift=.88in] {\begin{tabular}{@{}c@{}}Scores\\ $\hat{w}\in\R^L$\end{tabular}};
\node (ranking) [box, fill=lblue!30, right of=weights, xshift=.88in] {\begin{tabular}{@{}c@{}}Ranking\\operation $\mathcal{R}$\end{tabular}};
\node (output) [right of=ranking, xshift=.8in] {\begin{tabular}{@{}c@{}}Output\\(top-$k$ or\\ full ranking)\end{tabular}};
\node [below of=ranking, yshift=.1in] {(our work)};
\node [above of=alg, yshift=-.1in] {Step 1};
\node [above of=ranking, yshift=-.1in] {Step 2};
\path [draw, -latex'] (data) -- (alg);
\path [draw, -latex'] (alg) -- (weights);
\path [draw, -latex'] (weights) -- (ranking);
\path [draw, -latex'] (ranking) -- (output);
\end{tikzpicture}
\end{center}

Although the scores $\hat w$ obtained in Step 1 are often stable to small perturbations in $\cd$ (as in the preference voting setting) or can potentially be stabilized by methods such as bagging \citep{soloff2024stability}, the final output 
of $\topk$ or $\argsort$  can still be highly susceptible to slight changes in the scores. 
In other words, the stability in the learned scores $\hat{w}$ does not necessarily imply the stability of the induced ranked order, as a ranking operation is intrinsically discontinuous and therefore unstable; the situation is similar for $\topk$. Indeed, when the scores in $\hat{w}$ are close to each other, even a small perturbation to $\hat{w}$ can lead to drastically different rankings.

In the literature on both full rankings and top-$k$ selection, some degree of separation in the population scores is necessary for identifiability. In this setting, one assumes that $\hat w$ is a noisy realization of a ``true'' set of scores $w^*$, and theoretical results guaranteeing exact recovery to the true ranking (according to $w^*$) assume a minimum separation between scores for different items---see, e.g.,
\citet{chen2022optimal, chen2022partial} and \citet{chen2019spectral}. However, such assumptions are frequently violated in real-world applications. For example, in the Netflix movie rating data, many of the top-rated films have nearly identical average scores, making it difficult to justify any assumption on the minimum separation.

Prior work by \citet{soloff2024building} 
studied a special case of top-$k$ selection with $k=1$ in the context of classification. They proposed the \textit{inflated argmax} operator, which returns a \emph{set-valued} output (i.e., a subset of the candidate items) to account for potential ambiguity in Step 2. In this work, we focus on developing 
stable versions of top-$k$ and ranking operations for Step 2 that return informative set-valued outputs applicable to general ranking problems. 
Specifically, in the context of the full ranking problem, our ranking operation would return a set of possible rankings of the $L$ items, where we may return more than one possible ranking if needed to accommodate ambiguities. In the context of the top-$k$ problem, our procedure would return a set of possible top-$k$ items, where the number of items in the set may exceed $k$ if needed. 
We briefly summarize our main contributions below.

\subsection{Our contributions}\label{sec:contributions}
We propose two new notions of algorithmic stability tailored to ranking problems: top-$k$ stability and full ranking stability. 
As illustrated in the flow chart above, we focus on developing operators to stabilize Step 2 in ranking problems. To this end, we propose the \emph{inflated top-$k$} and the \emph{inflated full ranking} operations, generalizing the inflated argmax operator of \cite{soloff2024building}. We derive assumption-free stability guarantees for these two operators (\cref{main_thm:inflated-topk,main_thm:inflated-ranking}), discuss their connections to the inflated argmax (\cref{sec:connection-tripod}), and show they return minimal sets whenever possible (\cref{prop:topk-optimality,prop:fullranking-optimality}). Subsequent experiments on both real and synthetic data confirm that our proposed methods exhibit stability while remaining informative.

\section{A unified framework for stable ranking}
We start by defining notions of stability that are meaningful in the context of the top-$k$ selection problem and the full ranking problem. As discussed above in \Cref{sec:problem-setting&prior-work}, in order to enable stability guarantees even when the data might lead to some uncertainty in the ranking, we need to allow for some ambiguity in the output: a procedure for solving the top-$k$ problem might need to return a set of size $>k$, and a procedure for solving the full ranking problem might need to return more than one possible ranking.

We begin with the top-$k$ problem, for any fixed $k\in[L]$. Let $\powerset_{\geq k}([L])$ denote the set of subsets of $[L]$ of size $\geq k$ (this is a subset of $\powerset([L])$, the power set of $[L]$). We will consider the stability of a function\footnote{In the diagram in \Cref{sec:problem-setting&prior-work}, the function $f$ corresponds to $\cR \circ \ca$.} \[f: \cup_{n\geq 1}\mathcal{Z}^n\to\powerset_{\geq k}([L]),\] which maps a dataset $\cd\in\mathcal{Z}^n$ (for any sample size $n$) to a subset $f(\cd)\subseteq[L]$, with $|f(\cd)|\geq k$.  
In particular, we expect to have $|f(\cd)|>k$ if the data exhibits some ambiguity in terms of identifying the top-$k$ items.
For any dataset $\cd\in\mathcal{Z}^n$, we write $\cd^{\setminus i}\in\mathcal{Z}^{n-1}$ to denote this same dataset with the $i$th point removed---that is, if $\cd = (Z_1,\dots,Z_n)$ then $\cd^{\setminus i} = (Z_1,\dots,Z_{i-1},Z_{i+1},\dots,Z_n)$.
\begin{definition}[Stability for top-$k$ selection]\label{def:ranking-stability-topk}
    We say that a function $f:\cup_{n\geq 1}\mathcal{Z}^n\to\powerset_{\geq k}([L])$ has top-$k$ stability $\delta$ at sample size $n$ if
    \[\frac{1}{n}\sum_{i=1}^n\mathbf{1}\left\{\left|f(\cd)\cap f(\cd^{\setminus i})\right| \geq k\right\} \ge 1- \delta\textnormal{ for all $\cd\in\mathcal{Z}^n$}.\]
\end{definition}
To explain this definition, we are requiring that the answer returned for a dataset $\cd$ and for $\cd^{\setminus i}$ should be consistent with each other (for most $i$) in the sense that their overlap should have size $\geq k$ (since we are seeking to identify the top-$k$ items). In the special case $k=1$, this definition coincides with the notion of \emph{selection stability} proposed by \citet{soloff2024building} for the argmax problem (i.e., the problem of identifying the top-ranked item).\footnote{In many applications, estimation algorithms and ranking algorithms often incorporate some form of randomization---for instance, a random initialization, or, random tie-breaking rules for the ranking procedure. The definitions and results of this paper can be extended naturally to incorporate randomized algorithms, but for conciseness we do not present these extensions here and only consider the deterministic case.}

Next, we turn to the full ranking problem. For this setting, we will consider functions of the form
\[f:\cup_{n\geq 1}\mathcal{Z}^n\to \powerset(\mathcal{S}_L),\] where $\mathcal{S}_L$ is the set of permutations of $[L]$, and $\powerset(\mathcal{S}_L)$ is the power set of $\mathcal{S}_L$. That is, given a dataset $\cd$ of any size, $f(\cd)$ returns a set of permutations, each corresponding to a possible ranking. If $|f(\cd)|=1$, this can be interpreted to mean that the data suggests a single ranking, while $|f(\cd)|>1$ indicates some ambiguity.
\begin{definition}[Stability for full ranking]\label{def:ranking-stability-fullR}
We say that a function $f:\cup_{n\geq 1}\mathcal{Z}^n\to\powerset(\mathcal{S}_L)$ has full ranking stability $\delta$ at sample size $n$ if
    \[\frac{1}{n}\sum_{i=1}^n\mathbf{1}\left\{f(\cd)\cap f(\cd^{\setminus i}) \neq\varnothing\right\} \ge 1- \delta\textnormal{ for all $\cd\in\mathcal{Z}^n$}.\]
\end{definition}
In other words, the sets of possible rankings $f(\cd)$ and $f(\cd^{\setminus i})$ should be consistent with each other (for most $i$), in the sense that at least one permutation (i.e., one possible ranking of the $L$ items) should appear in both sets.

\subsection{Stability of the scores, or stability of the ranking?}
From this point on, we will focus our attention on procedures that follow the two-stage structure described in \Cref{sec:problem-setting&prior-work}, as is common across many applications. Specifically, we will consider functions of the form $f = \mathcal{R}\circ\ca$, where $\ca:\cup_{n\geq 1}\mathcal{Z}^n\to \R^L$ is a learning algorithm mapping a dataset $\cd$ to a vector of scores $\hat{w}=\ca(\cd)\in\R^L$, and we then apply a ranking procedure $\mathcal{R}$ to produce either an output $\mathcal{R}(\hat{w})\in\powerset_{\geq k}([L])$ (for the top-$k$ problem) or $\mathcal{R}(\hat{w})\in\powerset(\mathcal{S}_L)$ (for the full ranking problem).

As mentioned in \Cref{sec:problem-setting&prior-work}, the learning algorithm $\ca$ used to learn the scores $\hat{w}$ tends to be stable in many cases, or in other cases, it is easy to modify $\ca$ to ensure stability. Formally, we say that $\ca$ has $(\eps,\delta)$-stability at sample size $n$ if, for all datasets $\cd\in\mathcal{Z}^n$,
\begin{equation}\label{eqn:score-stability}
    \frac{1}{n}\sum_{i=1}^n \mathbf{1}\left\{\| \hat{w} - \hat{w}^{\setminus i}\| \geq \eps\right\} \le \delta,
\end{equation}
where $\hat{w} = \ca(\cd)$ and $\hat{w}^{\setminus i} = \ca(\cd^{\setminus i})$, and where $\|\cdot\|$ is the usual Euclidean (i.e., $\ell_2$) norm on $\R^L$.

This notion of algorithmic stability shares close ties with the formulations appearing in the work of \citet{elisseeff2005stability, soloff2024bagging, soloff2024stability}; related definitions of stability have previously been considered in the context of generalization bounds and learnability for ranking problems by \citet{lan2008query,agarwal2009generalization,gao2013uniform}. We present below two scenarios where the stability condition \eqref{eqn:score-stability} on $\ca$ is achieved in an assumption-free way.

\begin{itemize}
    \item 
In the preference voting setting, each participant casts a vote among $L$ candidate items---that is, the data values are $Z_i\in[L]$, denoting the vote from the $i$th participant. The final ranking scores can then be obtained by simply counting the fraction of votes each item receives, $\hat{w}_\ell = \frac{1}{n}\sum_i \mathbf{1}\{Z_i=\ell\}$, which is $(\eps,\delta)$-stable with $\eps = \sqrt{2}/n$ and $\delta=0$.
    \item For general learning algorithms $\ca$ with bounded outputs, \citet{soloff2024stability} show that one can apply bagging (bootstrapping) with $\ca$ as a base algorithm to achieve assumption-free stability, with $\eps^2\delta \propto 1/n$.

\end{itemize}

These examples illustrate that, in a two-stage ranking procedure $\mathcal{R}\circ\ca$, stability of the learning algorithm $\ca$ is often easily achievable---but crucially, this does not necessarily translate to stability in the output ranking list, due to the discontinuity of the ranking operation. For example, in a scenario where participants are voting for their favorite item from two candidate items, suppose that Item 1 receives 51\% of the vote while Item 2 receives 49\%, but if the $i$th participant is removed from the vote count then this flips---that is, $\hat{w}=(0.51,0.49)$ while $\hat{w}^{\setminus i} = (0.49,0.51)$. Then the perturbation in the output of the learning algorithm $\ca$ is equal to $\| \hat{w} - \hat{w}^{\setminus i}\|$, which is small---but the outcome of a ranking procedure $\mathcal{R}$ applied to $\hat{w}$, versus to $\hat{w}^{\setminus i}$, may give completely different answers.

In light of this challenge, our task from this point on is the following: \begin{quote}We aim to develop a ranking operation $\mathcal{R}$ (for the top-$k$ problem or the full ranking problem) such that, when combined with a learning algorithm $\ca$ that is stable as in~\eqref{eqn:score-stability}, then the two-stage procedure $\mathcal{R}\circ\ca$ is guaranteed to satisfy the appropriate notion of stability (that is, top-$k$ stability as in \Cref{def:ranking-stability-topk}, or full ranking stability as in \Cref{def:ranking-stability-fullR}).\end{quote}

\subsection{The inflated top-\texorpdfstring{$k$}{k} method}\label{sec:inflated-topk}
We are now ready to define our proposed ranking procedure for the top-$k$ problem:
\begin{definition}[Inflated top-$k$]\label{def:inflated-topk}
    For any $k \in [L]$, any $w\in\R^L$, and any $\eps>0$, define the $\eps$-inflated top-$k$ ranking as    \begin{equation}\label{eqn:inflated_topk_defn}
    \stopk(w) := \left\{j\in[L] : \dist(w,C_j^{\eps, k}) < \eps\right\},
    \end{equation}
    where
    \[C_j^{\eps,k} = \left\{v \in \R^L: v_j \ge v_{(k+1)} + \eps/\sqrt{2} \right\}.\]
\end{definition}
Here for any set $C\subseteq\R^L$, we define $\dist(w,C)=\inf_{v\in C}\|w-v\|$, and for any vector $v=(v_1,\dots,v_L)\in\R^L$, we write $v_{(1)}\geq \dots \geq v_{(L)}$ to denote the order statistics of the values $v_1,\dots,v_L$.

For the special case $k=1$, the definition above coincides with the \emph{inflated argmax} proposed by \citet{soloff2024building}---that is, for $k=1$, $\stopk(w) = \sargmax(w) := \left\{j\in[L] : \dist(w,C_j^{\eps,1}) < \eps\right\}$.

To help interpret this definition, we observe that $C_j^{\eps, k}$ is the set of score vectors $v\in\R^L$ for which the $j$th entry $v_j$ is in the top-$k$ by a positive margin. We can also observe that $\stopk(w)\supseteq\topk(w)$ always holds---that is, the top-$k$ entries are always included in the inflated top-$k$. Moreover, the inflated top-$k$ is naturally permutation-invariant, satisfying
\[\pi(i) \in \stopk\big((w_1,\dots,w_L)\big) \ \Longleftrightarrow \ i\in\stopk\big((w_{\pi(1)},\dots,w_{\pi(L)})\big)\]
for any $w\in\R^L$ and any $\pi\in\mathcal{S}_L$---that is, permuting the entries of $w$ does not alter our assessment of which values $w_j$ may lie in the top-$k$.

\paragraph{Stability guarantee:}
Our first main result verifies that, in a two-stage procedure for top-$k$ ranking, we can obtain a stability guarantee by combining the inflated top-$k$ method with any stable learning algorithm $\ca$.
\begin{theorem}\label{main_thm:inflated-topk}
    Fix any $n\geq 2$ and any $k \in [L]$.
    Let $\ca$ be any  learning algorithm that has $(\eps,\delta)$-stability at sample size $n$ as in~\eqref{eqn:score-stability}. Then the two-stage procedure $\stopk \circ \ca$ has top-$k$ stability $\delta$ at sample size $n$, as defined in \Cref{def:ranking-stability-topk}.
\end{theorem}
The key idea behind the proof of this result is the following property of the inflated top-$k$: 
\begin{equation}\label{eqn:compatibility-topk}
\textnormal{For all $w,v\in\R^L$, if $\|w-v\|< \eps$ then }
    \left|\stopk(w)\cap \stopk(v)\right| \ge k.
\end{equation}
This property allows us to use the stability property of the learning algorithm $\ca$~\eqref{eqn:score-stability} to derive top-$k$ stability for the two-stage procedure $\stopk\circ\ca$.

\paragraph{Optimality of the method:}
Since the inflated top-$k$ permits returning a larger set to handle ambiguity (that is, $\stopk(w)$ may contain $>k$ elements), a natural concern is whether this leads to excessive redundancy. Is the method returning overly large sets more often than necessary? The following proposition shows that there is no need for such concern: in fact, the inflated top-$k$ is optimal for the problem of stable top-$k$ selection, in the sense that it returns exactly $k$ elements as often as possible.
\begin{prop}\label{prop:topk-optimality}
Consider any function $\mathcal{R}:\R^L\to\powerset_{\geq k}([L])$. Suppose that $\mathcal{R}$ is permutation invariant, that
$\topk(w) \subseteq \mathcal{R}(w)$ for all $w\in\R^L$, and that $\mathcal{R}$ satisfies
\[\textnormal{For all $w,v\in\R^L$, if $\|w-v\|< \eps$ then }
    \left|\mathcal{R}(w)\cap \mathcal{R}(v)\right| \ge k.\]
Then for any $w\in\R^L$,
\[\textnormal{If $|\mathcal{R}(w)|=k$ then $\stopk(w)=\mathcal{R}(w)$}\]
(and consequently $|\stopk(w)|=k$).
\end{prop}
In other words, as long as $\mathcal{R}$ satisfies the conditions needed for ensuring top-$k$ stability when combined with any stable learning algorithm $\ca$ (analogous to the property~\eqref{eqn:compatibility-topk} satisfied by $\stopk$), then the inflated top-$k$ method is able to return a set of size $k$ (i.e., no ambiguity in identifying the top-$k$ items) at least as often as $\mathcal{R}$.

\paragraph{Efficient computation:}
Although computing the set returned by the inflated top-$k$~\eqref{eqn:inflated_topk_defn} appears complicated at first glance, involving calculating distances to the set $C_j^{\eps,k}$, its computation can be drastically simplified. To present this result, for simplicity we will assume that we are computing $\stopk(w)$ for a score vector satisfying $w_1\geq \dots \geq w_L$, without loss of generality.
\begin{prop}\label{prop:inflated_topk_compute}
    Fix any $k \in [L]$, any $\eps>0$ and any $w \in \R^L$ with $w_1\geq \dots \geq w_L$. Then it holds that
    \[\stopk(w) = \big\{1,\dots,k-1\big\} \cup \Big\{ k-1 + j : j\in \sargmax\big((w_k,\dots,w_L)\big)\Big\}.\]
\end{prop}
In other words, computing the inflated top-$k$ for a vector of scores $w=(w_1,\dots,w_L)$ (with $w_1\geq \dots \geq w_L$) is only as hard as computing the inflated argmax for the subvector $(w_k,\dots,w_L)$---and this latter problem has a computationally simple solution \citep[Prop.~11]{soloff2024building}.
Moreover, this result offers an interesting perspective on how the inflated top-$k$ method operates: if we think of the inflated argmax as handling any ambiguity at the argmax boundary (i.e., identifying which item is in the top position), then the inflated top-$k$ can be viewed as analogously handling any ambiguity at the boundary between the $k$-th and $(k+1)$-st position.

\subsection{The inflated full ranking method}\label{sec:inflated-ranking}
Next, we turn to the full ranking problem.
In a canonical sorting algorithm, the full ranking list can be constructed sequentially: at each step, the algorithm selects the argmax over the current set of elements, assigns it to the next position in the list, and then repeats this process on the reduced set until all positions are filled. Our inflated full ranking method follows a similar iterative strategy.
\begin{definition}[Inflated full ranking]\label{def:inflated-ranking}
For any $w\in\R^L$ and any $\eps>0$, define the $\eps$-inflated full ranking as
\begin{equation}\label{eqn:inflated-ranking} \sranking
    (w) :=  \left\{\pi \in \mathcal{S}_{L}: 1 \in \sargmax\big((w_{\pi(k)},\dots,w_{\pi(L)})\big) \ \textnormal{ for each } k \in [L] \right\}.
    \end{equation}
\end{definition}
In other words, for a permutation $\pi$ to be included into the set $\sranking(w)$, the following must hold: at each step $k\in[L]$, when we examine the subvector $(w_{\pi(k)},\dots,w_{\pi(L)})$, the inflated argmax must include its first entry (i.e., the entry corresponding to item $\pi(k)$ in the original vector).

We can immediately verify that the inflated full ranking satisfies several natural properties, by construction. First, it must hold that $\sranking(w)\ni \ranking(w)$---for instance, if $w_1\geq \dots \geq w_L$, then the identity permutation, $\pi = \textnormal{Id}$, must be included in the set of possible rankings $\sranking(w)$. Next, the procedure satisfies permutation invariance: for any $\pi\in\mathcal{S}_L$,
\[\pi\in\sranking\big((w_1,\dots,w_L)\big) \ \Longleftarrow \ \textnormal{Id} \in\sranking\big((w_{\pi(1)},\dots,w_{\pi(L)})\big)\]

\paragraph{Stability guarantee:}
As for the top-$k$ setting, our next result verifies that, when running a two-stage procedure for full ranking, we obtain a stability guarantee by combining the inflated full ranking method with any stable learning algorithm $\ca$.
\begin{theorem}\label{main_thm:inflated-ranking}
    Fix any $n\geq 2$.
    Let $\ca$ be any  learning algorithm that has $(\eps,\delta)$-stability at sample size $n$ as in~\eqref{eqn:score-stability}. Then the two-stage procedure $\sranking \circ \ca$ has full ranking stability $\delta$ at sample size $n$, as defined in \Cref{def:ranking-stability-fullR}.
\end{theorem}
As for the stability guarantee for the top-$k$ problem, here the key step will again be to verify that inflated full ranking satisfies the following property:
\begin{equation}\label{eqn:compatibility-fullR}
        \textnormal{For all $w,v\in\R^L$, if $\|w-v\|< \eps$ then }
        \sranking(w)\cap \sranking(v) \ne \varnothing.
\end{equation}

\paragraph{Optimality of the method:}
Next, we will examine the question of optimality: essentially, does the inflated full ranking return the smallest possible set of candidate permutations? The following result states that the method is optimal for the problem of stable full ranking in this sense.
\begin{prop}\label{prop:fullranking-optimality}
Consider any function $\mathcal{R}:\R^L\to\powerset(\mathcal{S}_L)$. Suppose that $\mathcal{R}$ is permutation invariant, that
$\ranking(w) \in \mathcal{R}(w)$ for all $w\in\R^L$, and that $\mathcal{R}$ satisfies
\[\textnormal{For all $w,v\in\R^L$, if $\|w-v\|< \eps$ then }
    \mathcal{R}(w)\cap \mathcal{R}(v) \ne \varnothing.\]
Then for any $w\in\R^L$, and any $i\neq j\in[L]$,
\[\textnormal{If $\pi^{-1}(i)<\pi^{-1}(j)$ for all $\pi\in\mathcal{R}(w)$, then $\pi^{-1}(i)<\pi^{-1}(j)$ for all $\pi\in\sranking(w)$.}\]
\end{prop}
In other words, if $\mathcal{R}(w)$ is able to claim that item $i$ is definitely ranked higher than item $j$ (based on score vector $w$), then the same is true for $\sranking(w)$.
In particular, this implies that if $\mathcal{R}(w)=\{\pi\}$ (i.e., $\mathcal{R}(w)$ returns a single permutation), then $\sranking(w)=\{\pi\}$ as well. But this result has additional implications as well. For example, suppose there are many near-ties within $w$, but there is a clear separation between the top-$k$ items and the remaining $L-k$ items---without loss of generality, suppose that $w_1,\dots,w_k$ are each ranked higher than any of $w_{k+1},\dots,w_L$. If this separation is unambiguous according to the ranking rule $\mathcal{R}$ (i.e., for every $\pi\in\mathcal{R}(w)$, it holds that $\{\pi(1),\dots,\pi(k)\}=\{1,\dots,k\}$), then the same is true for $\sranking(w)$.

\paragraph{Efficient computation:}
Since calculating the inflated argmax is computationally efficient (as discussed above), checking whether a given permutation $\pi\in\mathcal{S}_L$ lies in $\sranking(w)$ is simple---but, if $L$ is large, enumerating all such permutations $\pi$ may be computationally challenging. To help with this task, the following result shows that we can restrict our attention to a much smaller set of possible permutations $\pi$.
\begin{prop}\label{prop:inflated-full-ranking-restrict-range}
    For any $w\in\R^L$ and any $\eps>0$, it holds that
    \[\min\big\{k : j\in \stopk(w)\big\} \leq \pi^{-1}(j) \leq \sum_{\ell=1}^L \mathbf{1}\{w_\ell > w_j - \eps/\sqrt{2}\},\]
    for all $j\in[L]$ and all $\pi\in\sranking(w)$.
\end{prop}
That is, these upper and lower bounds restrict the possible positions assigned to the $j$th item by any permutation $\pi$ in the inflated full ranking.

\subsection{Connecting the full ranking, top-\texorpdfstring{$k$}{k}, and argmax problems}\label{sec:connection-tripod}
In the setting of standard ranking rules (without inflation), the argmax, top-$k$, and full ranking questions are all closely connected: for example, if $\pi = \ranking(w)$ determines the ranking of $w$, then $\pi(1)$ determines the argmax of $w$, and the top elements of $\pi$ determine the answer to the top-$k$ problem, i.e., $\topk(w) = \{\pi(1),\dots,\pi(k)\}$. Conversely, the ranking of $w$ can be uniquely determined by considering $\topk(w)$ across all values of $k$.

In this section, we extend these connections to the setting of the inflated top-$k$ and inflated full ranking methods. Indeed, as we have seen previously, the inflated argmax is a key ingredient in computing the inflated top-$k$ (\Cref{prop:inflated_topk_compute}), and in defining the inflated full ranking (\Cref{def:inflated-ranking}). Here, we examine some additional connections.

Our first result shows that the inflated full ranking of $w$ directly reveals the inflated top-$k$ set.
\begin{prop}\label{prop:from-fullrank-to-topk}
For any $\eps>0$, any $w\in\R^L$, and any $k\in[L]$,
\[\stopk(w) = \cup_{\pi\in\sranking(w)} \{\pi(1),\dots,\pi(k)\}.\]
\end{prop}
In other words, the inflated top-$k$ set consists of all items $j$ that appear in the top $k$ entries of any permutation $\pi\in\sranking(w)$. 

In the reverse direction, the picture is less straightforward. It turns out that computing the inflated top-$k$ sets (across all $k$) is not sufficient to reveal the inflated full ranking; the inflated full ranking contains information beyond what is revealed by the inflated top-$k$ sets. 
\begin{prop}\label{prop:from-topk-to-fullrank}
For any $\eps>0$ and any $w\in\R^L$, 
define 
\[\mathcal{R}(w) = \left\{ \pi\in\mathcal{S}_L : \pi(k) \in\stopk(w) \ \forall \ k\in[L]\right\}.\]
Then
$\sranking(w)\subseteq\mathcal{R}(w)$,
and moreover, there exist examples where this set inclusion is strict, i.e., $\sranking(w)\subsetneq\mathcal{R}(w)$.
\end{prop}

\section{Experiments}\label{sec:experiment}
In this section, we evaluate our proposed methods on real and simulated data.\footnote{Running the  experiments took approximately $30$ minutes on a MacBook Pro laptop using a single CPU core.
Code to reproduce the experiments is available at \url{https://github.com/jake-soloff/stability-ranking-experiments}.}

\subsection{Experiments for top-\texorpdfstring{$k$}{k} selection}\label{sec:experiment-topk}
\paragraph{Data and scores calculation.} We use the Netflix Prize data \citep{bennett2007netflix},\footnote{Data was obtained from \url{https://www.kaggle.com/datasets/netflix-inc/netflix-prize-data/data}.} which consists of $480,189$ anonymous Netflix customers' ratings over a total of $L=17,770$ movies. The ratings are on a scale from $1$ to $5$ stars. For each trial, we first generate a subsample of $n=1000$ users, sampled uniformly without replacement; let $Z_i$ denote the ratings data for each user $i\in[n]$. The score vector $\hat{w}=\ca(Z_1,\dots,Z_n)$ is defined by taking $\hat{w}_\ell$ to be the average score for the $\ell$th movie, averaged over all users who provided a rating for that movie, and modified with a ``+1'' term in the denominator for shrinkage, $\hat{w}_\ell = \frac{\textnormal{sum of all ratings for the $\ell$th movie}}{1+\textnormal{number of ratings for the $\ell$th movie}}$, which allows $\hat{w}_\ell$ to be well-defined even if a movie received no ratings among the $n$ sampled users. We then repeat this procedure for $N=100$ independent trials, i.e., each trial uses a new subsample of $n$ individuals.

\paragraph{Methods and evaluation.}
We compare the inflated top-$k$ against the usual (uninflated) top-$k$ selection method, with $k=20$ and with $\eps=0.01$ for the inflated method, under several metrics.\footnote{For top-$k$, if movies are tied for the $k$th position, we break ties by choosing the movie with the smallest index $\ell\in\{1,\dots,L\}$. Notably, another reason for the ``+1'' in the denominator is that, without this shrinkage term, we often see many ties between movies near the top of the ranked list (namely, movies that received only a few ratings, and all those ratings are equal to the highest value $5$, which often occurs due to the small subsample size $n$)---and since characterizing the instability of top-$k$ selection in the presence of frequent ties is heavily dependent on our choice of tie-breaking rule, to avoid this issue we instead use the shrinkage term.} 

For each trial $j=1,\dots,N$, let $\hat{w}^{(j)}\in\R^L$ denote the score vector obtained by computing the average rating (modified with the ``+1'' term as described above) over the users included in the $j$th trial, as described above, and let $\hat{w}^{(j),\backslash i}$ denote the same score vector when computed without the $i$th user in the subsample, for each $i=1,\dots,n$.
To assess the top-$k$ stability of each method, we compute, for each trial  $j=1,\ldots,N$, 
\[\delta_j = \frac{1}{n}\sum_{i=1}^n \mathbf{1}\left\{\left|\mathcal{R}(\hat{w}^{(j)}) \cap \mathcal{R}(\hat{w}^{(j),\setminus i})\right| < k \right\},\]
where $\mathcal{R}(\cdot)$ denotes either $\topk(\cdot)$ or $\stopk(\cdot)$.
(Recalling \Cref{def:ranking-stability-topk}, we should see $\delta_j\leq \delta$ for any method that satisfies top-$k$ stability at level $\delta$.)
We also consider an alternative measure to assess stability, the Jaccard similarity \citep{Jaccard}, which measures  overlap between sets; a value closer to $1$ suggests stronger agreement (i.e., higher stability). For each trial $j$, and each of the two methods, we compute
\[
    \mathrm{Jaccard}_j = \frac{1}{n}\sum_{i=1}^{n} \frac{\left|\mathcal{R}(\hat{w}^{(j)}) \cap \mathcal{R}(\hat{w}^{(j),\setminus i})\right|}{\left|\mathcal{R}(\hat{w}^{(j)}) \cup \mathcal{R}(\hat{w}^{(j),\setminus i})\right|}.
\]
Finally, we also compute the size of the returned set,
$\textnormal{Size}_j =\left|\mathcal{R}(\hat{w}^{(j)})\right|$,
which indicates the informativeness of the set; ideally we would want to return exactly $k$ items, but a set size larger than $k$ indicates that there may be ambiguity in the scores, with near-ties between multiple movies.

\begin{table}
    \centering
    \renewcommand{\arraystretch}{1.6}
    \begin{tabular}{c|c|c|c|c}
    \rowcolor{gray!20}
        Methods & $\max_{j\in [N]}\delta_j$ &  $\frac{1}{N}\sum_{j \in [N]}\delta_j$ & $\frac{1}{N}\sum_{j \in [N]}\mathrm{Jaccard}_j$ & $\frac{1}{N}\sum_{j \in [N]}\textnormal{Size}_j$  \\
        \hline
        $\topk$ & $0.8530$ & $0.1205 \ (0.0136)$ & $0.9876 \ (0.0015)$ & $20.00 \ (0.0000)$  \\
        \hline
        $\stopk$ & $0.0380$ & $0.0094 \ (0.0009)$ &  
        $0.9906 \ (0.0006)$ & 
        $21.22 \ (0.1101)$ \\
    \end{tabular}
    \caption{Results on the Netflix Prize dataset (see \Cref{sec:experiment-topk} for details). Evaluation results under various metrics are reported in the table, with standard errors for the averages shown in parentheses.}
    \label{tab:top-k20}
\end{table}

\paragraph{Results.}
\Cref{tab:top-k20} reports the results of the experiment in terms of the evaluation metrics defined above. First, we observe that the inflated top-$k$ shows much better stability than top-$k$, with substantially lower values of $\delta_j$ (and with slightly higher Jaccard similarity), on average across the $N$ trials. Note that even $\max_{j\in[N]}\delta_j$ is quite small for inflated top-$k$, agreeing with our theoretical finding that inflated top-$k$ offers distribution-free stability, i.e., uniformly over any data set.  
The high instability of top-$k$ suggests the presence of  ambiguity within the data, which makes the returned set of top-$k$ highly sensitive to the removal of a single data point.
The inflated top-$k$ accommodates this ambiguity by including slightly more possible candidates in the list. There is relatively little cost to this, since the size of the set returned by the inflated top-$k$ is only slightly larger, returning (on average) a set of size $\approx 21.22$, as compared to $k=20$. See \Cref{app:netflix} for figures illustrating the results, and additional results at different values of $k$. 

\subsection{Experiments for full ranking}\label{sec:experiment-fullR}

\paragraph{Data and scores calculation.}
We generate data from a Gaussian linear model, $Y_i = X_i^\top \beta^* + \zeta_i$, where $\zeta_i\stackrel{\textnormal{iid}}{\sim}\mathcal{N}(0, 1)$ and where the feature vectors $X_i\in\R^L$ are drawn as $X_i\stackrel{\textnormal{iid}}{\sim} \mathcal{N}(0, \Sigma)$, for $\Sigma_{ij} = \rho^{|i-j|}$. We set $n=50$, $L=5$, and $\rho=0.5$. The regression coefficients are taken to be $\beta^*_j = j/\sqrt{\sum_{k=1}^Lk^{2}}$. The learning algorithm $\mathcal{A}$ is defined as follows. First we estimate the coefficients of the regression via $\ell_2$-constrained least squares: writing $Z_i = (X_i,Y_i)$ for the $i$th data point, define \[\hat\beta = \mathrm{argmin}
_{\|\beta\|_2\le 1} \sum_{i=1}^n (Y_i - X_i^\top \beta)^2.\]
(The stability of algorithms of this type (in the sense of~\eqref{eqn:score-stability}) has been well-studied---see, e.g., \cite{bousquet2002stability}.) We then take $\hat{w}_\ell = |\hat\beta_\ell|$ to estimate the magnitude of the $\ell$th coefficient---our goal will now be to rank the coefficients---that is, to rank the features in terms of their (estimated) importance in the model.

\paragraph{Methods and evaluation.}
We compare inflated full ranking (with $\eps=0.05$) against the usual full ranking procedure, under several metrics: for each $j\in[N]$, we compute
\[\delta_j = \frac{1}{n}\sum_{i=1}^{n}\mathbf{1}\left\{\calr(\hat{w}^{(j)})\cap \calr(\hat{w}^{(j),\setminus i}) = \varnothing\right\},\ \ \textnormal{Size}_j =\big|\calr(\hat{w}^{(j)})\big|,\]
for $\mathcal{R}(\cdot)$ denoting either $\ranking(\cdot)$ or $\sranking(\cdot)$, to assess the stability and the informativeness of each procedure.

\paragraph{Results.} We present results for this simulation in \Cref{table:sim}. The inflated full ranking procedure shows substantially better stability, with much smaller values of $\delta_j$. This comes at little cost, because the inflated full ranking procedure returns $\approx 1.75$ many permutations on average---we can interpret this as saying that there is ambiguity among only very few of the coefficients.

\begin{table}
    \centering
    \renewcommand{\arraystretch}{1.6}
    \begin{tabular}{c|c|c|c}
    \rowcolor{gray!20}
        Methods & $\max_{j\in [N]}\delta_j$ &  $\frac{1}{N}\sum_{j \in [N]}\delta_j$ & $\frac{1}{N}\sum_{j \in [N]}\textnormal{Size}_j$\\
        \hline
        $\ranking$ & $0.78$ & $0.1757 \ (0.0049)$ & $1.00 \ (0.00)$ \\
        \hline
        $\sranking$ & $0.12$ & $ 0.0162 \ (0.0007)   $ & $1.76 \ (0.04)$  
    \end{tabular}
    
    \caption{Results on full ranking stability for simulated data (see \Cref{sec:experiment-fullR} for details). Evaluation results under various metrics are reported in the table, with standard errors for the averages shown in parentheses.}\label{table:sim}
\end{table}

\section{Discussion}\label{sec:discussion}

In this section, we describe related works on the problem of learning a ranking, and their connections to our proposed stable ranking. We then summarize our main contributions, and discuss some limitations and potential extensions that provide interesting avenues for future work.

\subsection{Related work}
\citet{devic2024stability} also propose a method that can reflect the uncertainty of the learned score predictors from Step 1. However, their work differs from ours in considering score \textit{distributions} for each item (not point scores) and returning a probabilistic distribution over ranking lists, whereas we aim for deterministic set-valued outputs. Notably, the notion of a stochastic type of output to account for the uncertainty in ranking problems has long been employed, though most works focus on promoting fairness instead of algorithmic stability. See, for example \citet{dwork2019learning, singh2021fairness, guo2023inference}.
\citet{oh2024finest} apply fine-tuning to empirically stabilize a given recommendation system, focusing on sequential recommendation based on users' previous interactions, which is distinct from our setting. 
Another related line of work focuses on proposing measures of ranking stability, along with corresponding empirical evaluations \citep{adomavicius2016classification,asudeh2018obtaining, oh2022rank}.

Our definition of ranking stability is rooted in the perspective of using set-valued output to quantify uncertainty, which connects to set-valued classification \citep{grycko1993classification,del2009learning,lei2014classification, chzhen2021set} and conformal prediction \citep{vovk2005algorithmic,sadinle2019least,angelopoulos2023recommendation,angelopoulos2023conformal}. Interestingly, \citet{guo2023inference} also use this set-valued idea to implicitly encode information about ranking scores' separation when constructing their probabilistic distribution over ranking lists, though they primarily aim to address the fairness problem.

\subsection{Summary and future directions}
This work describes novel notions of stability for two ranking problems: identifying the top-$k$ items among a collection of candidates, and ranking all candidates. Our framework seeks to ensure that the removal of a sample from the training dataset will not yield a wholesale change in the returned collection of items (for top-$k$) or permutations of the items (for full rankings). More specifically, in the context of top-$k$ selection, our method returns a set of $k$ \textit{or more} items, and is stable in the sense that removing one sample and rerunning our method typically yields a set that shares at least $k$ items with the original output. Similarly, in the context of full ranking, our method returns a set of permutations (rankings) of the $L$ items, and is stable in the sense that removing one sample and rerunning our method typically yields a set of permutations with at least one in common with the original output. Of course, these particular definitions offer one specific notion of what it means to be stable in the context of a ranking problem, and exploring other possible formulations of stability is an important open question.

Our approach does not rely upon assumptions on the base algorithm $\ca$ used to assign scores to each item based on the training dataset $\cd$, and does not place any distributional assumptions on $\cd$. Our guarantees hold for any dataset size $n\geq 2$. Furthermore, the returned sets of items or permutations are optimal in that they are as small as possible, maximizing the informativeness of the returned top-$k$ selections or full rankings. 
Our stability guarantees depend on the algorithm $\ca$ mapping a dataset $\cd$ to a vector of scores for each of the $L$ items being $(\varepsilon,\delta)$-stable. Bagging can be used to ensure this property \citep{soloff2024stability}, but whether alternative methods can provide similar assumption-free guarantees with less computational complexity remains an open question for further work.

\subsection*{Acknowledgements}
The authors gratefully acknowledge the National Science Foundation via grant DMS-2023109, and the support of the NSF-Simons AI-Institute for the Sky (SkAI) via grants NSF AST-2421845 and Simons Foundation MPS-AI-00010513. J.S. and R.F.B. were partially supported by the Office of Naval Research via grant N00014-24-1-2544. J.S. was also partially supported by the Margot and Tom Pritzker Foundation. R.M.W. was partially supported by the NSF-Simons National Institute for Theory and Mathematics in Biology (NITMB) via grants NSF (DMS-2235451) and Simons Foundation (MP-TMPS-00005320).

\bibliographystyle{apalike} 
\bibliography{citations}

\appendix

\section{Proofs of theoretical results}

\subsection{Proof of \Cref{main_thm:inflated-topk}}
As discussed after the statement of the theorem, it is sufficient to verify 
\[\textnormal{For all $w,v\in\R^L$, if $\|w-v\|< \eps$ then }
    \left|\stopk(w)\cap \stopk(v)\right| \ge k,\]
since assuming this property holds, we then have
\[\frac{1}{n}\sum_{i=1}^n\mathbf{1}\left\{\left|\stopk(\ca(\cd))\cap \stopk(\ca(\cd^{\setminus i}))\right| \geq k\right\} \ge\frac{1}{n}\sum_{i=1}^n\mathbf{1}\left\{\|\ca(\cd) - \ca(\cd^{\setminus i})\|<\eps\right\} \ge 1- \delta.\]

To prove that the above property holds, we will need to use the fact that the same property holds for the inflated argmax, i.e., for the case $k=1$ \citep[Thm.~9]{soloff2024building}:
\begin{equation}\label{eqn:sargmax-eps-compatible}\textnormal{For all $w,v\in\R^L$, if $\|w-v\|< \eps$ then }\sargmax(w)\cap\sargmax(v)\neq \varnothing.\end{equation}
We will also need the following lemma:
\begin{lemma}\label{lem:topk-as-union}
    For any $\eps>0$, any $w\in\R^L$, and any $k\in[L]$,
    \[\stopk(w) = \bigcup_{\substack{ L-k+1 \leq \ell \leq L, \\ \textnormal{distinct }i_1,\dots,i_\ell\in[L]}} \left\{i_j : j\in\sargmax\big((w_{i_1},\dots,w_{i_\ell})\big)\right\}.\]
\end{lemma}
That is, the inflated top-$k$ set is given by all entries $w_j$ that are selected by the inflated argmax, when we apply the inflated argmax to any subvector of size $\geq L-k+1$.

Now fix any $w,v\in\R^L$ with $\|w-v\|<\eps$. 
Let $S=\stopk(w)\cap\stopk(v)$. Suppose that $|S|<k$. Let $i_1,\dots,i_{L-|S|}\in[L]$ enumerate the remaining indices, $[L]\backslash S$. Then
\[\left\|\big(w_{i_1},\dots,w_{i_{L-|S|}}\big) - \big(v_{i_1},\dots,v_{i_{L-|S|}}\big)\right\|\leq \|w-v\| < \eps,\]
and so by~\eqref{eqn:sargmax-eps-compatible},
\[\sargmax\big((w_{i_1},\dots,w_{i_{L-|S|}})\big) \cap \sargmax\big((v_{i_1},\dots,v_{i_{L-|S|}})\big)\neq\varnothing.\]
We can therefore find some $j$ with
\[j\in \sargmax\big((w_{i_1},\dots,w_{i_{L-|S|}})\big) \cap \sargmax\big((v_{i_1},\dots,v_{i_{L-|S|}})\big).\]
But since $L-|S| > L-k$, by \Cref{lem:topk-as-union} this means that $i_j\in\stopk(w)$ and $i_j\in\stopk(v)$. Since $\stopk(w)\cap\stopk(v)=S\not\ni i_j$, we have reached a contradiction, which completes the proof.

\subsection{Proof of \Cref{prop:topk-optimality}}
Without loss of generality, assume $w_1\geq \dots \geq w_L$. Then $ \{1,\dots,k\}= \topk(w) \subseteq\mathcal{R}(w)$ by assumption, and so $|\mathcal{R}(w)|=k$ implies $\mathcal{R}(w) = \{1,\dots,k\}$.
Now fix any $\ell>k$ and define
\[ v= (w_1,\dots,w_{k-1},w_\ell,w_{k+1},\dots,w_{\ell-1},w_k,w_{\ell+1},\dots,w_L),\]
which is simply the vector $w$ with $k$th and $\ell$th entries swapped.
By permutation invariance of $\mathcal{R}$, this means that $\mathcal{R}(v) = \{1,\dots,k-1,\ell\}$, and consequently $|\mathcal{R}(w)\cap\mathcal{R}(v)| = k-1<k$. 
Therefore,
\[\eps \leq \|w-v\| = \sqrt{2} |w_k - w_\ell|.\]
Since $w_k\geq w_\ell$ by assumption, this means that 
\[w_k \geq w_\ell + \eps/\sqrt{2}.\]

This holds for every $\ell>k$, i.e., we have shown that
\[w_k \geq \max\{w_{k+1},\dots,w_L\} + \eps/\sqrt{2}.\]
By \citet[Lemma 15]{soloff2024building}, this implies that
\[\sargmax((w_k,\dots,w_L)) = \{1\},\]
i.e., the inflated argmax (applied to the subvector $(w_k,\dots,w_L)$) returns a singleton set. By \Cref{prop:inflated_topk_compute}, we have
\[\stopk(w) = \{1,\dots,k-1\} \cup\{k-1 + j : j\in\sargmax((w_k,\dots,w_L))\} = \{1,\dots,k-1\}\cup\{k\},\]
which completes the proof.

\subsection{Proof of \Cref{prop:inflated_topk_compute}}
First we will show that 
\[\stopk(w) \subseteq \big\{1,\dots,k-1\big\} \cup \Big\{ k-1 + j : j\in \sargmax\big((w_k,\dots,w_L)\big)\Big\}.\]
Fix any $j\in\stopk(w)$. Since the right-hand side above must include all indices $j\leq k$, we can assume $j>k$ to avoid the trivial case. By definition of $\stopk(w)$, we can find some $v\in C_j^{\eps,k}$ with $\|w-v\|<\eps$.

Next, let $v^{-j}_{(1)} \geq \dots \geq v^{-j}_{(L-1)}$ be the order statistics of $(v_i)_{i\neq j}$, and define
\[\tilde{v} = \big(v^{-j}_{(1)}, \dots, v^{-j}_{(j-1)}, v_j, v^{-j}_{(j)},\dots,v^{-j}_{(L-1)}\big)\in\R^L.\]
Since $w_1 \geq \dots \geq w_L$, by the rearrangement inequality we have
\begin{align*}
    \|w - \tilde{v}\|^2
    &=(w_j - v_j)^2 + \left\| (w_1,\dots,w_{j-1},w_{j+1},\dots,w_L) - \big(v^{-j}_{(1)}, \dots, v^{-j}_{(j-1)}, v^{-j}_{(j)},\dots,v^{-j}_{(L-1)}\big) \right\|^2\\
    &\leq (w_j - v_j)^2 + \left\| (w_1,\dots,w_{j-1},w_{j+1},\dots,w_L) - (v_1,\dots,v_{j-1},v_{j+1},\dots,v_L)\right\|^2\\
    &=\|w-v\|^2 < \eps^2.
\end{align*}
Moreover, $\tilde{v}_{(k+1)} = v_{(k+1)}$ (since $\tilde{v}$ is simply a permutation of $v$), and therefore we have
\[\tilde{v}_j = v_j \geq v_{(k+1)} + \eps/\sqrt{2} = v^{-j}_{(k)} + \eps/\sqrt{2} = \max_{\ell\geq k, \ell\neq j} \tilde{v}_\ell + \eps/\sqrt{2},\]
where the inequality holds since $v\in C_j^{\eps,k}$, and the following step holds since we clearly must have $v_j$ in the top $k$ elements of $v$, and so $v_{(k+1)} = v^{-j}_{(k)}$. 
Finally, by \citet[Lemma~15]{soloff2024building}, for any vector $u$, $\sargmax(u)=\{j\}$ if and only if $u \in C^{\eps,1}_j$. Consequently, 
\[\sargmax\big((\tilde{v}_k,\dots,\tilde{v}_L)\big) = \{j - k+1\},\]
i.e., the unique element selected is the one corresponding to the entry $\tilde{v}_j = v_j$.

Next, we calculate
\[\big\|(w_k,\dots,w_L) - (\tilde{v}_k,\dots,\tilde{v}_L)\big\|\leq \|w-\tilde{v}\|<\eps.\]
Therefore, $\sargmax\big((w_k,\dots,w_L)\big)\cap\sargmax\big((\tilde{v}_k,\dots,\tilde{v}_L)\big)\neq \varnothing$ by~\eqref{eqn:sargmax-eps-compatible}, which now implies $j-k+1 \in \sargmax((w_k,\dots,w_L))$. This completes the first part of the proof.

Next we need to show the converse, i.e.,
\[\stopk(w) \supseteq \big\{1,\dots,k-1\big\} \cup \Big\{ k-1 + j : j\in \sargmax\big((w_k,\dots,w_L)\big)\Big\}.\]
Since $\stopk(w)\supseteq\topk(w)$ by construction, we only need to consider any $j\in \sargmax\big((w_k,\dots,w_L))$.
If this holds for some $j$, then by definition of the inflated argmax, there must be some vector $v\in\R^{L-k+1}$ with 
\[\|(w_k,\dots,w_L) - v\|<\eps, \ v_j \geq \max_{i\neq j} v_i + \eps/\sqrt{2}.\]
Now define
\[\tilde{v} = (w_1,\dots,w_{k-1},v_1,\dots,v_{L-k+1}).\]
Then clearly,
\[\|w - \tilde{v}\| = \|(w_k,\dots,w_L) - v\|<\eps.\]
Moreover, 
\[\tilde{v}_{k-1+j} = v_j \geq \max_{i\neq j} v_i + \eps/\sqrt{2}
= \max_{\ell \geq k, \ell \neq k-1+j} \tilde{v}_\ell + \eps/\sqrt{2}
\geq \tilde{v}_{(k+1)} + \eps/\sqrt{2},\]
and therefore $\tilde{v}\in C_{k-1+j}^{\eps,k}$. Consequently, $k-1+j\in\stopk(w)$ by definition, which completes the proof.

\subsection{Proof of \Cref{main_thm:inflated-ranking}}

As discussed after the statement of the theorem, it is sufficient to verify 
\[\textnormal{For all $w,v\in\R^L$, if $\|w-v\|< \eps$ then }
    \sranking(w)\cap\sranking(v)\neq \varnothing,\]
since assuming this property holds, we then have
\begin{multline*}\frac{1}{n}\sum_{i=1}^n\mathbf{1}\left\{\sranking(\ca(\cd))\cap \sranking(\ca(\cd^{\setminus i}))\neq \varnothing\right\}\\ \ge\frac{1}{n}\sum_{i=1}^n\mathbf{1}\left\{\|\ca(\cd) - \ca(\cd^{\setminus i})\|<\eps\right\} \ge 1- \delta.\end{multline*}

Fix any $w,v\in\R^L$ with $\|w-v\|<\eps$. We will now iteratively construct a permutation $\pi\in \sranking(w)\cap\sranking(v)$.

First, since $\|w-v\|<\eps$, it holds that $\sargmax(w)\cap\sargmax(v)\neq\varnothing$, by~\eqref{eqn:sargmax-eps-compatible}. We can therefore choose $\pi(1)$ to be any index
\[\pi(1) \in \sargmax(w)\cap\sargmax(v).\]
Next we proceed by induction. Suppose that we have defined $\pi(1),\dots,\pi(k-1)$, and are now ready to define $\pi(k)$. Let $S_k = [L]\backslash \{\pi(1),\dots,\pi(k-1)\}$. Then
\[\|w_{S_k} - v_{S_k}\| \leq \|w-v\|<\eps,\]
and so $\sargmax(w_{S_k})\cap\sargmax(v_{S_k})\neq\varnothing$, by~\eqref{eqn:sargmax-eps-compatible}. In particular, we can choose some $j\in\sargmax(w_{S_k})\cap\sargmax(v_{S_k})$. Now let $\pi(k)\in {S_k}$ be chosen such that $\pi(k)$ corresponds to the $j$th element of $S_k$.

Proceeding iteratively as above, we have defined $\pi\in\mathcal{S}_L$. Now we verify that $\pi\in\sranking(w)\cap\sranking(v)$. For each $k$, note that $S_k$ is equal to some permutation of $\{\pi(k),\dots,\pi(L)\}$. By the permutation invariance of the inflated argmax, and by definition of $\pi(k)$, we therefore have
\[1\in\sargmax\big((w_{\pi(k)},\dots,w_{\pi(L)})\big).\]
Since this holds for every $k$, we have shown that $\pi\in\sranking(w)$. The same argument holds for $v$ as well, which completes the proof.

\subsection{Proof of \Cref{prop:fullranking-optimality}}

Without loss of generality, assume $w_1\geq \dots \geq w_L$. Then $\textnormal{Id} = \ranking(w) \in\mathcal{R}(w)$ by assumption. Since we assume $\pi^{-1}(i)<\pi^{-1}(j)$ for all $\pi\in\mathcal{R}(w)$, this means that we must have $i<j$. Next define
\[ v= (w_1,\dots,w_{i-1},w_j,w_{i+1},\dots,w_{j-1},w_i,w_{j+1},\dots,w_L),\]
which is simply the vector $w$ with $i$th and $j$th entries swapped. By permutation invariance of $\mathcal{R}$, this means that $\pi^{-1}(i) > \pi^{-1}(j)$ for all $\pi\in\mathcal{R}(v)$. Consequently $\mathcal{R}(w)\cap\mathcal{R}(v)=\varnothing$, and therefore
\[\eps \leq \|w-v\| = \sqrt{2} |w_i-w_j|.\]
Since $w_i\geq w_j$ by assumption, this means that 
\[w_i \geq w_j + \eps/\sqrt{2}.\]

Next fix any $\pi\in\sranking(w)$. Let $\pi^{-1}(i)=k$ and $\pi^{-1}(j)=\ell$. By definition of $\sranking(w)$, we have
\[1 \in \sargmax\big((w_{\pi(\ell)},\dots,w_{\pi(L)})\big).\]
By \citet[Prop.~20]{soloff2024building}, for any vector $u$, $\sargmax(u)\subseteq\{i : u_i > \max_{i'} u_{i'} - \eps/\sqrt{2}\}$, and therefore this means that
\[w_{\pi(\ell)} > \max_{\ell'\geq  \ell}w_{\pi(\ell')} - \eps/\sqrt{2}.\]
On the other hand, from our work above we know that
\[w_{\pi(k)} = w_i \geq w_j + \eps/\sqrt{2} = w_{\pi(\ell)} + \eps/\sqrt{2}.\]
This proves that we cannot have $k\geq \ell$---and consequently, $\pi^{-1}(i) < \pi^{-1}(j)$, as desired.

\subsection{Proof of \Cref{prop:inflated-full-ranking-restrict-range}}
Suppose $\pi^{-1}(j)=k$.
First, by \Cref{prop:from-fullrank-to-topk}, it holds that $j = \pi(k)\in\stopk(w)$. This establishes the lower bound. 

Next, by definition of the inflated full ranking, for every $\ell \leq k$ we have
\[1 \in \sargmax\big((w_{\pi(\ell)},\dots,w_{\pi(k)},\dots,w_{\pi(L)})\big).\]
By \citet[Prop.~20]{soloff2024building}, for any vector $v$, $\sargmax(v)\subseteq\{i : v_i > \max_{i'} v_{i'} - \eps/\sqrt{2}\}$, and consequently, we have
\[w_{\pi(\ell)} > \max_{\ell'=\ell,\dots,L}w_{\pi(\ell')} - \eps/\sqrt{2} \geq w_{\pi(k)} - \eps/\sqrt{2} = w_j - \eps/\sqrt{2}.\]
Since this holds for every $\ell\leq k$, we therefore have
\[\sum_{\ell=1}^L \mathbf{1}\{w_\ell > w_j - \eps/\sqrt{2}\}
= \sum_{\ell=1}^L \mathbf{1}\{w_{\pi(\ell)} > w_j - \eps/\sqrt{2}\}
\geq k,
\]
which completes the proof of the upper bound.

\subsection{Proof of \Cref{prop:from-fullrank-to-topk}}
First, fix any $\pi\in\sranking(w)$ and any $j\leq k$. Then $1\in\sargmax((w_{\pi(j)},\dots,w_{\pi(L)}))$, by definition of the inflated full ranking. Defining $i_1 = \pi(j),\dots,i_{L-j+1} = \pi(L)$, and applying \Cref{lem:topk-as-union}, we see that $\pi(j) \in \stopk(w)$. This proves that
\[\stopk(w)\supseteq \cup_{\pi\in\sranking(w)} \{\pi(1),\dots,\pi(k)\} .\]

Now we prove the converse. Without loss of generality assume $w_1\geq \dots \geq w_L$. Fix any $j\in\stopk(w)$. If $j\leq k$, then $j\in\{\pi(1),\dots,\pi(k)\}$ for $\pi=\textnormal{Id}$, which satisfies $\pi=\ranking(w)\in\sranking(w)$. If instead $j>k$ then define \[\pi = (1,\dots,k-1,j,k,\dots,j-1,j+1,\dots,L),\]
that is, the permutation $\pi$ places $w_j$ into position $k$ and otherwise sorts the entries of $w$ from largest to smallest, so that $j\in\{\pi(1),\dots,\pi(k)\}$.
For each $\ell\neq k$, we have $1 \in\sargmax((w_{\pi(\ell)},\dots,w_{\pi(L)}))$, since the first entry of this subvector is its maximum. For $\ell=k$, we have $j-k+1 \in\sargmax((w_k,\dots,w_L))$, by \Cref{prop:inflated_topk_compute}. By permutation invariance, then, $1\in\sargmax((w_j,w_k,\dots,w_{j-1},w_{j+1},\dots,w_L))$. This verifies that $\pi\in\sranking(w)$.
We have therefore proved that
\[\stopk(w)\subseteq \cup_{\pi\in\sranking(w)} \{\pi(1),\dots,\pi(k)\} .\]

\subsection{Proof of \Cref{prop:from-topk-to-fullrank}}
First fix any $\pi\in\sranking(w)$. Then by definition, for each $k\in[L]$ we have
\[1\in\sargmax((w_{\pi(k)},\dots,w_{\pi(L)})).\]
By \Cref{lem:topk-as-union}, this means that $\pi(k)\in\stopk(w)$. Since this holds for all $k$, we have proved that $\pi\in\mathcal{R}(w)$---and therefore, $\sranking(w)\subseteq\mathcal{R}(w)$.

Next, consider the following example: let $L=3$, $\eps = 1$, and 
\[w = ( 1 , 0.5 , 0 ).\]
Then we can calculate
\[\stopk(w) = \begin{cases} \{1,2\}, & k=1,\\
\{1,2,3\}, & k=2,\\
\{1,2,3\}, & k=3.\end{cases}\]
Choosing $\pi = (2,3,1)$, we therefore see that $\pi\in\mathcal{R}(w)$. However, 
\[\sranking(w) = \big\{ (1,2,3), (2,1,3), (1,3,2)\} \not\ni \pi.\]

\subsection{Proof of \Cref{lem:topk-as-union}}
Without loss of generality, assume $w_1 \geq \dots \geq w_L$. First, by \Cref{prop:inflated_topk_compute}, we have
\[\stopk(w) = \big\{1,\dots,k-1\big\} \cup \Big\{ k-1 + j : j\in \sargmax\big((w_k,\dots,w_L)\big)\Big\}.\]
Now fix any $\ell\in\stopk(w)$. If $\ell\leq k-1$, then let $i_1 = \ell, \dots, i_{L-\ell+1}=L$. Then \[1\in\sargmax((w_{i_1},\dots,w_{i_{L-\ell+1}})),\] since the first entry of this subvector is the largest. 
If instead $\ell\geq k$, then we must have
\[\ell = k-1 + j\textnormal{ where }j\in \sargmax\big((w_k,\dots,w_L)\big).\]
Now let $i_1 = k,\dots,i_{L-k+1} = L$. Then
\[j\in\sargmax((w_{i_1},\dots,w_{i_L})).\]
Combining these cases, we have proved that 
\[\stopk(w) \subseteq \bigcup_{\substack{ L-k+1 \leq \ell \leq L, \\ \textnormal{distinct }i_1,\dots,i_\ell\in[L]}} \left\{i_j : j\in\sargmax\big((w_{i_1},\dots,w_{i_\ell})\big)\right\}.\]

Now we prove the converse. Fix any distinct $i_1,\dots,i_\ell$ for $\ell\geq L-k+1$, and suppose $j\in \sargmax\big((w_{i_1},\dots,w_{i_\ell})\big)$. We now need to show that $i_j\in\stopk(w)$. By definition of the inflated argmax, there is some vector $v\in\R^\ell$, with $v_j\geq \max_{i\neq j} v_i + \eps/\sqrt{2}$, such that 
\[\|(w_{i_1},\dots,w_{i_\ell}) - v\|<\eps.\]
Now define $\tilde{v}\in\R^L$ with entries
\[\tilde{v}_{i_1} = v_1, \ \dots, 
 \tilde{v}_{i_\ell}=v_\ell,\]
and $\tilde{v}_i=w_i$ for all $i\not\in\{i_1,\dots,i_\ell\}$. Then
\[\|w - \tilde{v}\| = \|(w_{i_1},\dots,w_{i_\ell}) - v\|<\eps.\]
Moreover, 
\begin{multline*}\tilde{v}_{i_j} = v_j \geq \max\{v_1,\dots,v_{j-1},v_{j+1},\dots,v_\ell\} + \eps/\sqrt{2} \\
= \max\{\tilde{v}_{i_1},\dots,\tilde{v}_{i_{j-1}},\tilde{v}_{i_{j+1}},\dots,\tilde{v}_{i_\ell}\} + \eps/\sqrt{2} \geq \tilde{v}_{(L-\ell+2)}  + \eps/\sqrt{2} \geq \tilde{v}_{(k+1)}  + \eps/\sqrt{2} .
\end{multline*}
Therefore, $\tilde{v}\in C_{i_j}^{\eps,k}$, and consequently we have $i_j\in\stopk(w)$, as desired. This verifies that
\[\stopk(w) \supseteq \bigcup_{\substack{ L-k+1 \leq \ell \leq L, \\ \textnormal{distinct }i_1,\dots,i_\ell\in[L]}} \left\{i_j : j\in\sargmax\big((w_{i_1},\dots,w_{i_\ell})\big)\right\},\]
which completes the proof.

\section{Additional experiments}\label{app:netflix}
Under the same settings as in \Cref{sec:experiment-topk}, here we present additional experiment results for different values of $k$. We also plot the empirical distribution of $\delta_j$, across trials $j=1,\dots,N$, for each value of $k$.
Overall, we observe qualitatively similar results across the different values of $k$, although both methods are more stable for smaller values of $k$ (since there is less ambiguity among the top few movies, than for larger $k$).

\begin{table}[hb]
    \centering
    \renewcommand{\arraystretch}{1.6}
    \begin{tabular}{c|c|c|c|c|c}
    \rowcolor{gray!20}
        $k$ & Methods &$\max_{j\in [N]}\delta_j$ &  $\frac{1}{N}\sum_{j \in [N]}\delta_j$ & $\frac{1}{N}\sum_{j \in [N]}\mathrm{Jaccard}_j$ & $\frac{1}{N}\sum_{j \in [N]}\textnormal{Size}_j$  \\
        \hline
        \multirow{2}{*}{$k=5$} &$\topk$ & $0.5350$ & $0.0222 \ (0.0057)$ & $0.9926 \ (0.0019)$ & $5.00 \ (0.0000)$  \\
        &$\stopk$ & $0.0210$ & $0.0036 \ (0.0004)$ & $0.9925 \ (0.0010)$ & $5.33 \ (0.0567)$ \\
        \hline
        \multirow{2}{*}{$k=10$} &$\topk$ & $0.6560$ & $0.0828 \ (0.0125)$ & $0.9847 \ (0.0023)$ & $10.00 \ (0.0000)$  \\
        &$\stopk$ & $0.0300$ & $0.0059 \ (0.0006)$ & $0.9911 \ (0.0008)$ & $10.75 \ (0.0942)$ \\
        \hline
        \multirow{2}{*}{$k=20$} &$\topk$ & $0.8530$ & $0.1205 \ (0.0136)$ & $0.9876 \ (0.0015)$ & $20.00 \ (0.0000)$  \\
        &$\stopk$ & $0.0380$ & $0.0094 \ (0.0009)$ &  
        $0.9906 \ (0.0006)$ & 
        $21.22 \ (0.1101)$ \\
        \hline
        \multirow{2}{*}{$k=50$} &$\topk$ & $0.8940$ & $0.2666 \ (0.0230)$ & $0.9873 \ (0.0012)$ & $50.00 \ (0.0000)$  \\
        &$\stopk$ & $0.0800$ & $0.0135 \ (0.0016)$ & 
        $0.9922 \ (0.0003)$ & 
        $52.30 \ (0.1712)$ \\
        \hline
        \multirow{2}{*}{$k=100$}&$\topk$ &$0.9150$ & $0.3928$ \ $(0.0235)$ & $0.9893$ \ $(0.0008)$ & $100.00$ \ $(0.0000)$ \\
        & $\stopk$ & $0.0630$ & $0.0122$ \ $(0.0011)$ & $0.9939$ \ $(0.0003)$& $103.20$ \ $(0.1766)$\\
    \end{tabular}
    \caption{Results on the Netflix Prize dataset (see \Cref{sec:experiment-topk} for details). Evaluation results under various metrics are reported in the table, with standard errors for the averages shown in parentheses. (The data for $k=20$ is exactly as reported in \Cref{sec:experiment-topk}.)}
    \label{tab:top-k3510}
\end{table}

\begin{figure}
    \centering
    \begin{subfigure}{\textwidth}
        \centering
        \includegraphics[width=0.5\textwidth]{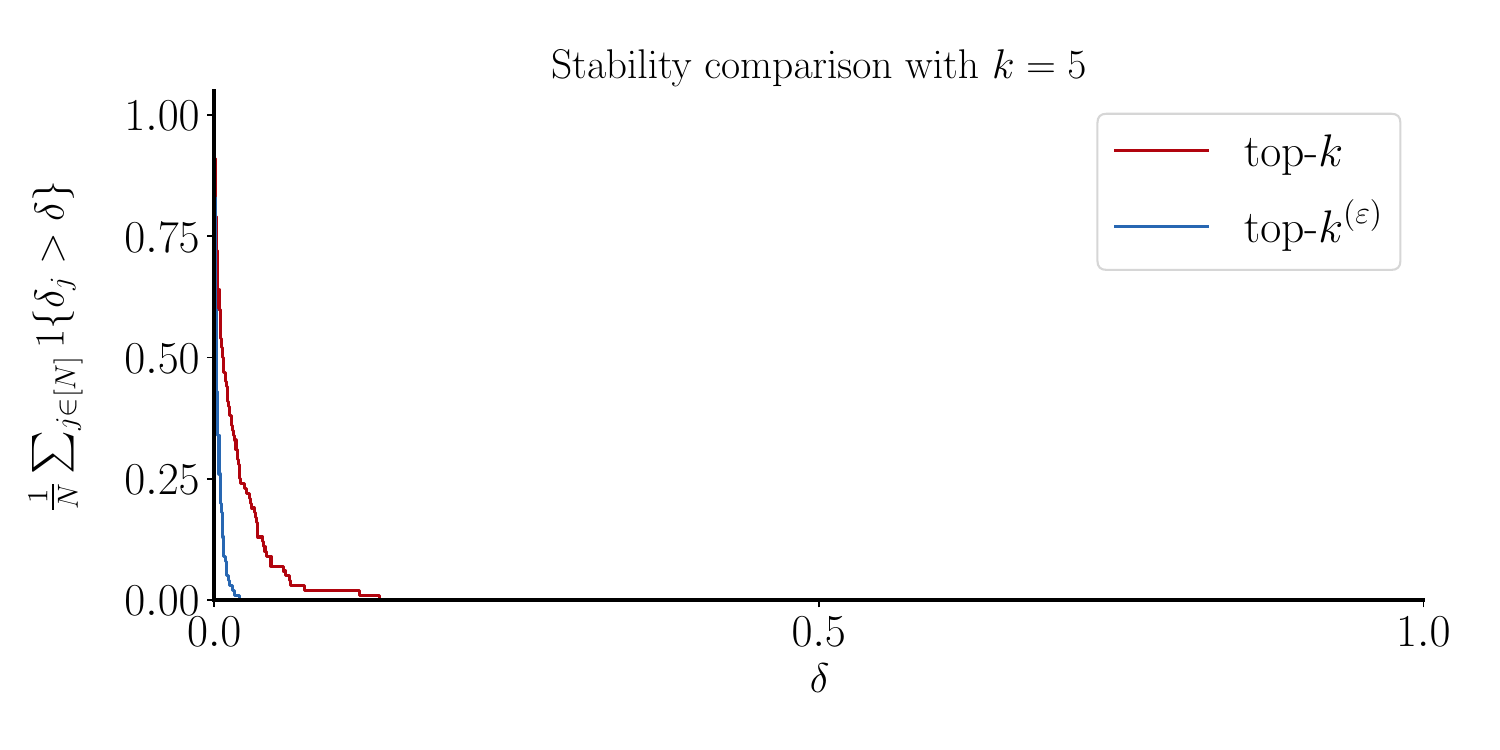}
    \end{subfigure}
    \vspace{1em}
    \begin{subfigure}{\textwidth}
        \centering
        \includegraphics[width=0.5\textwidth]{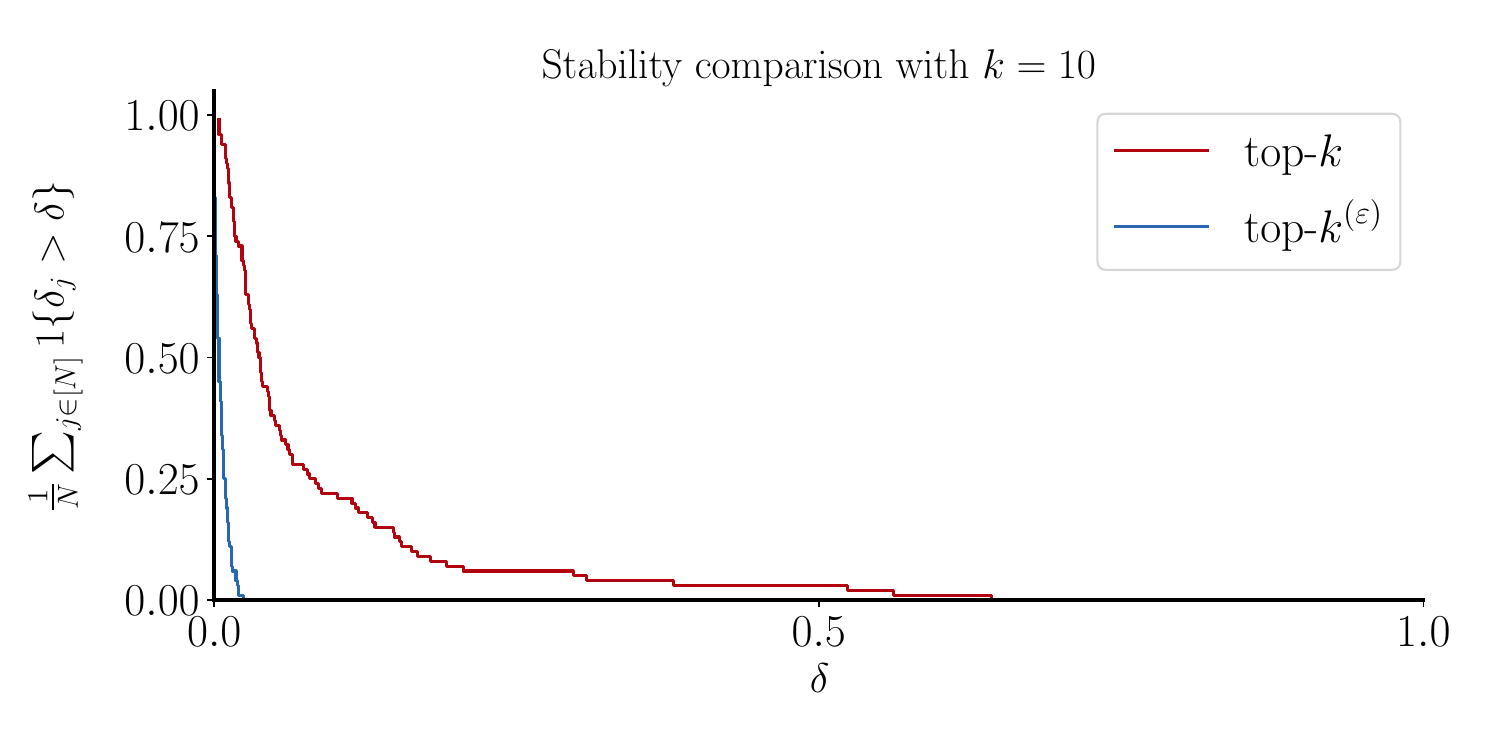}
    \end{subfigure}
    \vspace{1em}
    \begin{subfigure}{\textwidth}
        \centering
        \includegraphics[width=0.5\textwidth]{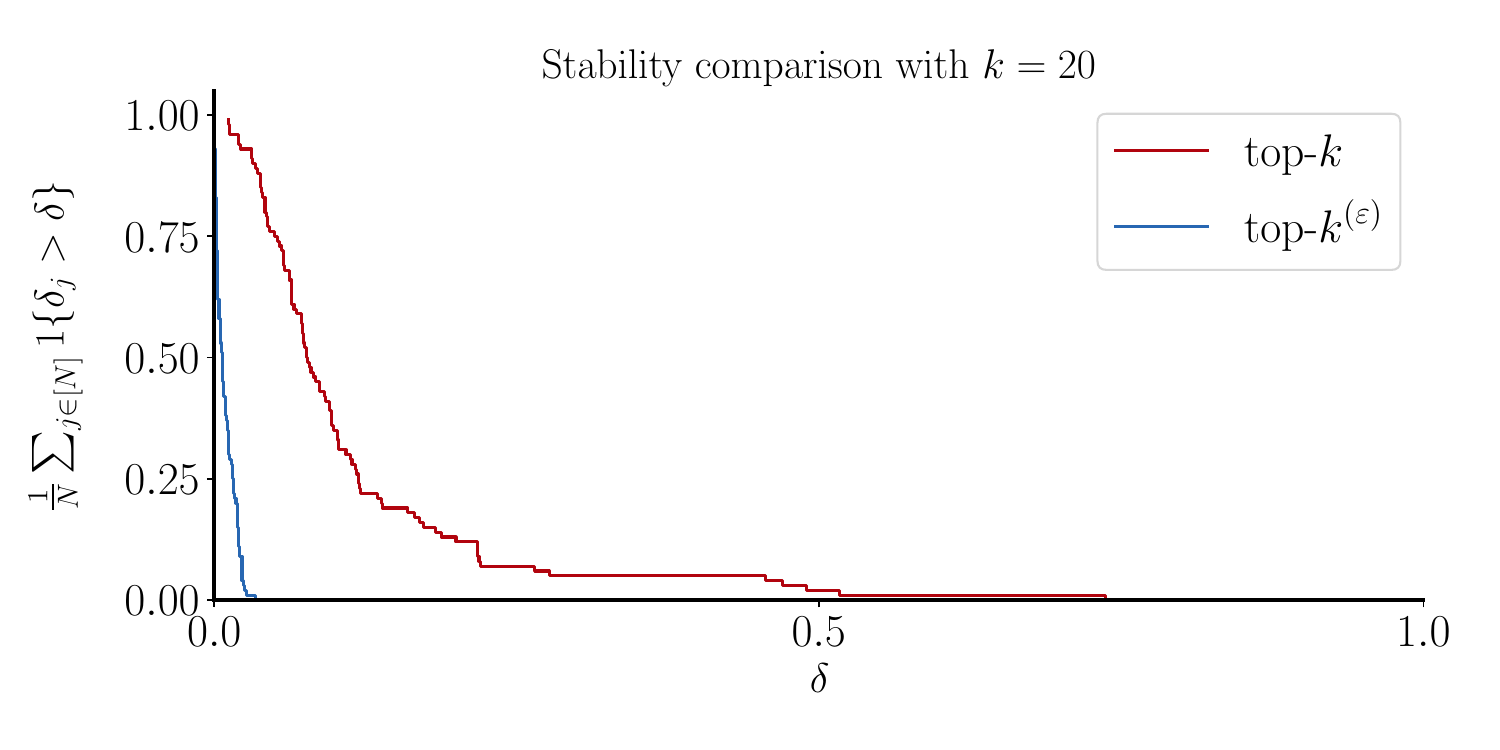}
    \end{subfigure}
     \vspace{1em}
        \begin{subfigure}{\textwidth}
        \centering
        \includegraphics[width=0.5\textwidth]{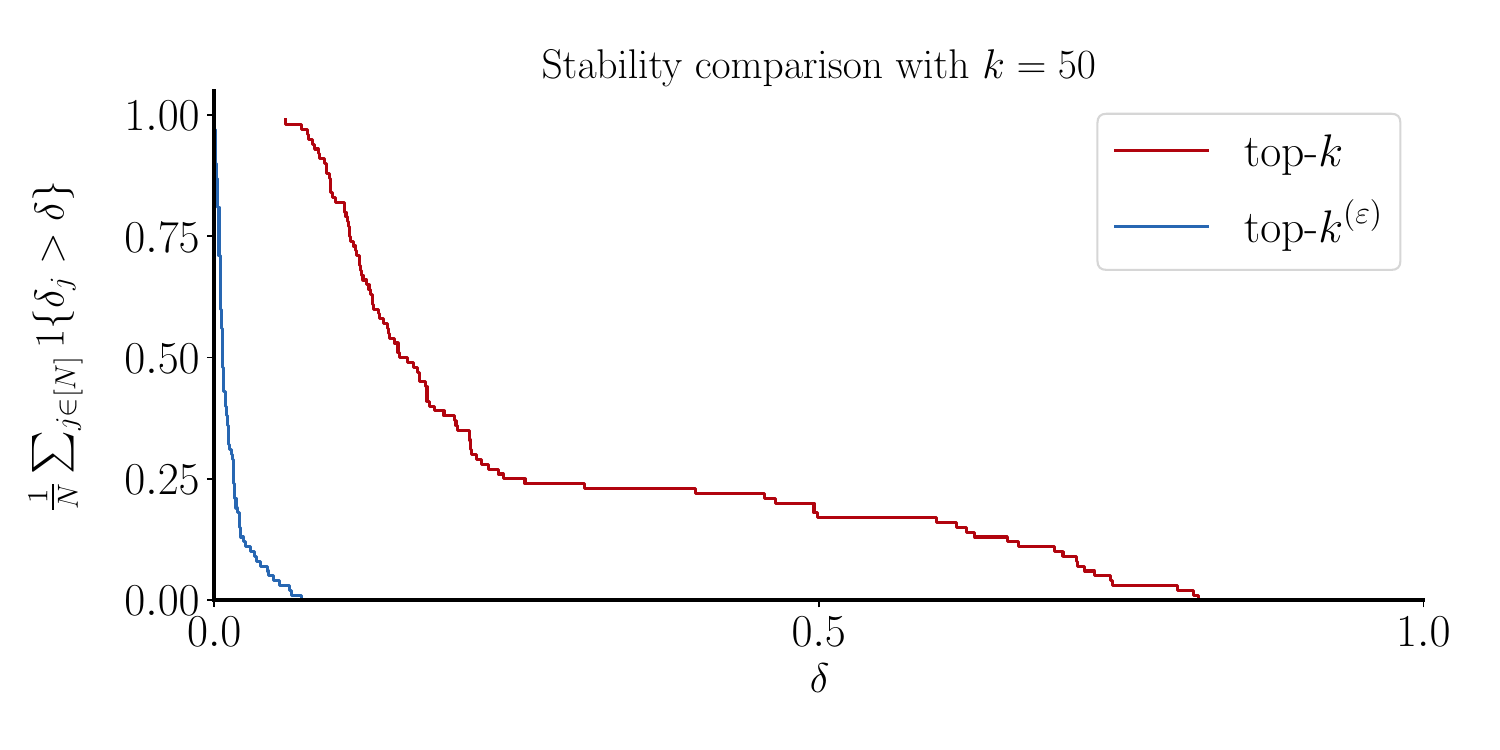}
    \end{subfigure}
    \vspace{1em}
        \begin{subfigure}{\textwidth}
        \centering
        \includegraphics[width=0.5\textwidth]{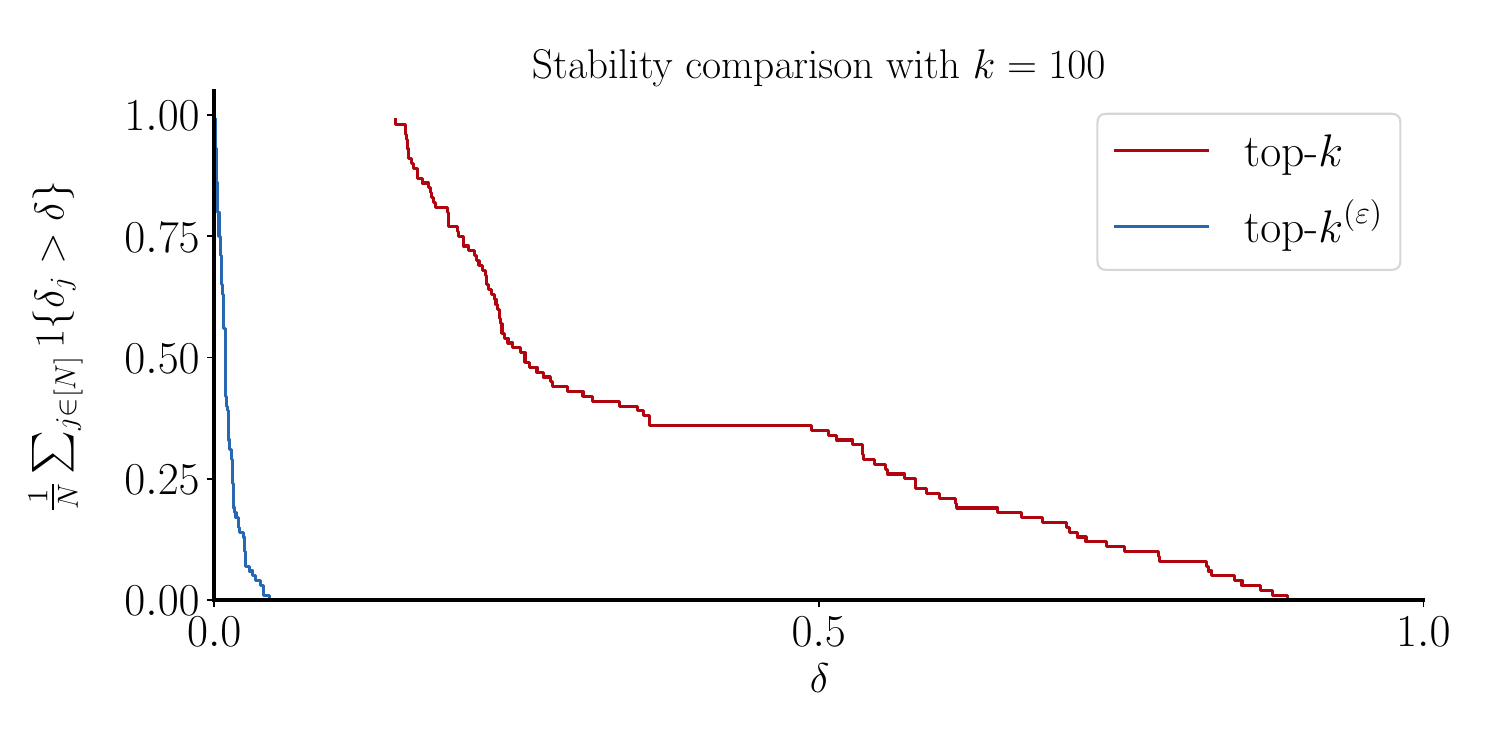}
    \end{subfigure}
    \caption{Results on the Netflix Prize dataset (see \Cref{sec:experiment-topk} for details). The plots show the distribution of $\delta_j$, across trials $j=1,\dots,N$, for each choice of $k$ and for each of the two methods.
    }
    \label{fig:topk-instability-diffk}
\end{figure}

\clearpage

\end{document}